\documentclass[conference]{IEEEtran}
\IEEEoverridecommandlockouts
\usepackage{cite}
\usepackage{amsmath,amssymb,amsfonts}
\usepackage{algorithmic}
\usepackage{graphicx}
\usepackage{textcomp}
\usepackage{xcolor}
\usepackage{tcolorbox} 
\usepackage{multirow}
\usepackage{colortbl}
\usepackage{pifont}
\usepackage{booktabs}
\newcommand{\eg}{\textit{e.g.}}

\def\BibTeX{{\rm B\kern-.05em{\sc i\kern-.025em b}\kern-.08em
    T\kern-.1667em\lower.7ex\hbox{E}\kern-.125emX}}
\begin{document}

\title{VisionNVS: Self-Supervised Inpainting for Novel View Synthesis under the Virtual-Shift Paradigm}


\author{
\textbf{Hongbo Lu}\textsuperscript{1,3,*},
\textbf{Liang Yao}\textsuperscript{2,*},
\textbf{Chenghao He}\textsuperscript{3},
\textbf{Fan Liu}\textsuperscript{2}\\
\textbf{Wenlong Liao}\textsuperscript{3},
\textbf{Tao He}\textsuperscript{3},
\textbf{Pai Peng}\textsuperscript{3,†}
\\
\textsuperscript{1}Shanghai Jiao Tong University \quad
\textsuperscript{2}Hohai University \quad
\textsuperscript{3}COWARobot Co. Ltd. \\
{\tt\small \textsuperscript{*}Equal Contribution \quad \textsuperscript{†}Corresponding Author
}
\\
}
\maketitle

\begin{abstract}

A fundamental bottleneck in Novel View Synthesis (NVS) for autonomous driving is the inherent supervision gap on novel trajectories: models are tasked with synthesizing unseen views during inference, yet lack ground truth images for these shifted poses during training. 
In this paper, we propose VisionNVS, a camera-only framework that fundamentally reformulates view synthesis from an ill-posed extrapolation problem into a self-supervised inpainting task. By introducing a ``Virtual-Shift'' strategy, we use monocular depth proxies to simulate occlusion patterns and map them onto the original view. This paradigm shift allows the use of raw, recorded images as pixel-perfect supervision, effectively eliminating the domain gap inherent in previous approaches. Furthermore, we address spatial consistency through a Pseudo-3D Seam Synthesis strategy, which integrates visual data from adjacent cameras during training to explicitly model real-world photometric discrepancies and calibration errors. 
Experiments demonstrate that VisionNVS achieves superior geometric fidelity and visual quality compared to LiDAR-dependent baselines, offering a robust solution for scalable driving simulation.
\end{abstract}

\begin{IEEEkeywords}
Autonomous driving, Novel View Synthesis, Inpainting
\end{IEEEkeywords}

\section{Introduction}
\label{sec:intro}
Photorealistic closed-loop simulation has become indispensable for the scalable validation of autonomous driving systems~\cite{zhao2025survey,cui2024survey,liu2025boost,yao2025remotesam}, particularly for testing safety-critical corner cases that are rarely encountered in the real world. To enable truly reactive simulation, Novel View Synthesis (NVS)~\cite{kong20253d4dworldmodeling} frameworks must possess the capability to synthesize high-fidelity camera views on free trajectories, allowing the virtual ego-vehicle to deviate laterally or longitudinally from its recorded path. While recent advances in Neural Radiance Fields (NeRF)~\cite{he2024neural,liao2025survey} and 3D Gaussian Splatting~\cite{bao20253d,fei20243d} have achieved remarkable results in static scene reconstruction, they often struggle to maintain geometric consistency and visual fidelity when extrapolating to unobserved viewpoints in unbounded, dynamic driving scenes. Generative approaches~\cite{yao2025remotereasoner} have emerged as a promising alternative by leveraging diffusion priors to hallucinate realistic details. However, applying them to the strict geometric constraints of the driving domain reveals a critical limitation in current methodologies.

A fundamental bottleneck in achieving robust free-trajectory NVS~\cite{guo2025dist,zhang2025toy,liao2024ef} is the inherent supervision gap between training and inference. Existing methods typically operate under an extrapolation paradigm. During training, they attempt to synthesize a target view from a novel pose, yet no ground truth image exists for this shifted perspective. To provide supervision, state-of-the-art approaches~\cite{guo2025dist,zhang2025toy}, such as FreeVS~\cite{wang2024freevs}, resort to warping geometric priors to the virtual viewpoint, creating ``pseudo-images" as training targets.
However, these methods typically rely on expensive LiDAR sensors to acquire accurate geometry, which limits their scalability to low-cost hardware platforms. When transitioning to a camera-only setting, one must resort to monocular depth estimation to construct geometric proxies. Unfortunately, projecting these estimated 3D points to a novel perspective inevitably exacerbates geometric artifacts. Due to the inherent inaccuracies of monocular depth (\eg, scale ambiguity and edge noise), the warped pseudo-targets are plagued by severe dislocation, far exceeding the noise levels of LiDAR-based projections.
Consequently, forcing a generative model to align with these defective pseudo-targets results in the propagation of artifacts, fundamentally limiting the visual quality and geometric fidelity of the synthesized views.

To surmount these challenges, we introduce VisionVNS, a purely vision-based framework that fundamentally reformulates view synthesis from an ill-posed extrapolation problem into a well-posed self-supervised inpainting task. 
Our key insight is that once the model learns how to fill in the black void regions, it will be able to reconstruct unknown areas in new viewpoints when the vehicle moves to a different perspective.
Leveraging this observation, we propose a Virtual-Shift strategy. Rather than warping features to a virtual pose, which introduces unavoidable artifacts, we simulate the occlusion patterns induced by a spatial shift and apply them as masks to the original viewpoint. This inversion allows us to utilize the raw, artifact-free recorded image as pixel-perfect ground truth, effectively eliminating the domain shift between the training condition and the supervision signal. Furthermore, to address the challenge of ``blind spots" where monocular context is insufficient, we incorporate a Pseudo-3D Seam Synthesis mechanism. By strategically filling masked regions with visual context from adjacent cameras during training, we explicitly force the model to internalize and resolve real-world photometric discrepancies and calibration noises. This ensures that the generated views remain spatially consistent even when extrapolating to trajectories far from the training distribution. 

Extensive experiments demonstrate the superiority of our framework. Remarkably, despite operating under a significantly more constrained setting, relying solely on monocular images without access to precise LiDAR geometry, our method achieves synthesis quality that rivals and often surpasses state-of-the-art methods. This performance validates that our self-supervised inpainting paradigm can effectively hallucinate geometrically correct details even with minimal sensor input.

Our contributions are summarized as follows:
\begin{itemize}
    \item We propose VisionVNS, a camera-only framework that reformulates NVS from ill-posed extrapolation to self-supervised inpainting, enabling pixel-perfect supervision without relying on LiDAR.

    \item We introduce the Virtual-Shift strategy and Pseudo-3D Seam Synthesis. These designs allow training on raw images as ground truth while explicitly robustifying the model against geometric occlusions and multi-view inconsistencies.

    \item Experiments on nuScenes demonstrate that VisionNVS surpasses LiDAR-dependent baselines in visual fidelity and geometric consistency, while enabling scalable streaming inference with constant memory overhead.
\end{itemize}

\section{Related Work}


\subsection{Neural Scene Reconstruction and Rendering}
To achieve photorealistic novel view synthesis (NVS), neural rendering techniques have been extensively adapted for driving scenes. NeRF-based methods, such as EmerNeRF~\cite{yang2023emernerf}, tackle the challenge of dynamic scene decomposition by learning separate static and dynamic fields via self-supervision. The emergence of 3D Gaussian Splatting (3DGS) has further revolutionized this domain due to its real-time rendering capabilities. StreetGaussian~\cite{huang2024s3gaussian} and PVG~\cite{chen2023periodic} explicitly model dynamic traffic elements, achieving high-fidelity reconstruction of moving vehicles. Despite their superior visual quality, these reconstruction-based methods typically rely on per-scene optimization, requiring computationally expensive training for each new video sequence. This lack of generalization capability severely limits their scalability for large-scale data generation and closed-loop simulation.

\subsection{Generalizable and Feed-Forward View Synthesis}
Bridging the gap between video generation and 3D reconstruction, recent research focuses on feed-forward architectures that generalize across scenes without per-case optimization. FreeVS~\cite{wang2024freevs} proposes a fully generative approach using pseudo-image priors derived from LiDAR projections to synthesize view-consistent images on free trajectories. Similarly, DriveForward~\cite{tian2025drivingforward} introduces a feed-forward 3DGS model capable of reconstructing dynamic scenes from sparse surround-view inputs in real-time. Most notably, DiST-4D~\cite{guo2025dist} decouples temporal and spatial generation, utilizing metric depth as a geometric bridge to enable both future prediction and spatial NVS within a unified diffusion framework. Unlike purely reconstruction-based methods, these approaches leverage large-scale pre-training to achieve zero-shot generalization, representing the state-of-the-art in scalable 4D autonomous driving simulation.


\begin{figure*}[t]
    \centering
    \includegraphics[width=0.99\linewidth]{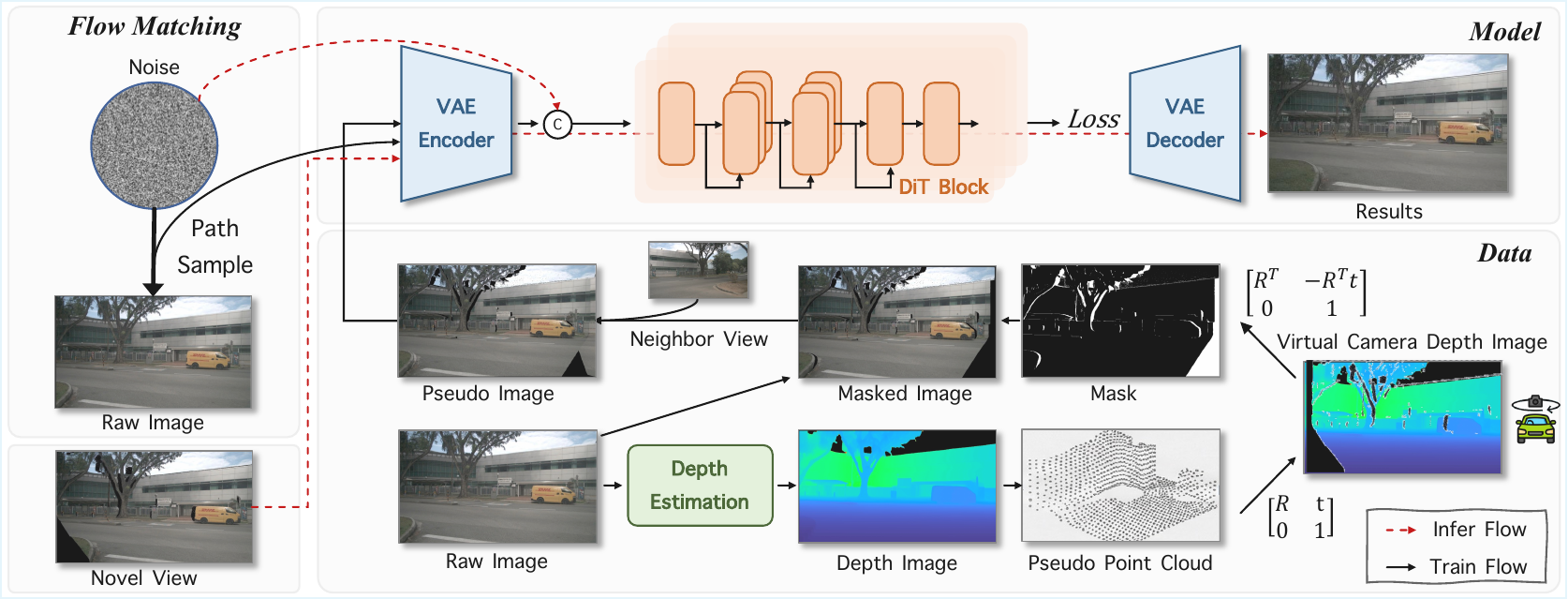}
    \caption{Overview of VisionNVS. Our camera-only framework reformulates novel view synthesis as self-supervised inpainting, trained via Flow Matching with a DiT backbone and the Wan2.2~\cite{wan2025wan} streaming VAE for efficient encoding. As shown in the bottom data pipeline, we construct a ``Pseudo Image" condition by masking the raw image based on virtual shift occlusions and filling gaps with neighbor views. The model (top) then learns to recover the original, artifact-free raw image from this degraded condition. 
    }
    \label{overview}
\end{figure*}

\section{Methodology}
\label{sec:method}

We propose \textbf{VisionNVS}, a camera-only framework that reformulates free-trajectory novel view synthesis from an ill-posed extrapolation problem into a self-supervised inpainting task, as shown in Fig.~\ref{overview}. Unlike existing methods that rely on LiDAR to project artifact-prone pseudo-targets, VisionNVS utilizes raw monocular videos as pixel-perfect supervision. 

\subsection{Problem Formulation}
\label{sec:formulation}

Standard NVS formulates the task as predicting a target view $\mathbf{I}_{tgt}$ given a source $\mathbf{I}_{src}$ and pose shift $\Delta \mathbf{T}$. Since ground truth for free trajectories is unavailable, prior works rely on a pseudo-target $\tilde{\mathbf{I}}_{tgt}$ generated via geometric projection. However, $\tilde{\mathbf{I}}_{tgt}$ inevitably suffers from projection artifacts (\eg, irregular black curves) due to monocular depth noise. Training with $\mathcal{L} = \|\hat{\mathbf{I}} - \tilde{\mathbf{I}}_{tgt}\|$ forces the model to overfit to these artifacts.

We propose to invert the problem into self-supervised inpainting. Instead of predicting a noisy virtual target, we aim to recover the artifact-free Raw Image $\mathbf{I}_{raw}$ from a degraded condition $\mathbf{I}_{cond}$. We construct the condition by simulating a virtual view shift:
\begin{equation}
    \mathbf{I}_{cond} = \mathbf{I}_{raw} \odot (1 - \mathbf{M}) + \mathbf{I}_{neighbor} \odot \mathbf{M},
\end{equation}
where $\mathbf{M}$ is the occlusion mask derived from the virtual projection (identifying lost pixels), and $\mathbf{I}_{neighbor}$ is the context synthesized from adjacent cameras. Crucially, our loss minimizes the distance to the perfect raw image:
\begin{equation}
    \min_{\theta} \mathbb{E} \left[ \| \mathcal{F}_{\theta}(\mathbf{I}_{cond}) - \mathbf{I}_{raw} \| \right].
\end{equation}
This ensures the model explicitly learns to resolve geometric and photometric inconsistencies without ever seeing defective ground truths.

\subsection{Monocular Geometric Proxies}
\label{sec:geometry}

Our framework adopts a camera-only setup, completely discarding active sensors like LiDAR. To obtain the necessary 3D information for view synthesis, we first lift the 2D raw images into 3D ego-centric space using monocular depth estimation.

Given a raw input image $\mathbf{I}_{raw} \in \mathbb{R}^{H \times W \times 3}$, we employ a pre-trained monocular depth estimator $\mathcal{E}_{depth}$ to predict a dense depth map $\mathbf{D}$. Using the camera intrinsic matrix $\mathbf{K}$, we back-project each pixel $\mathbf{u}=(u,v)$ into a 3D point $\mathbf{p}_{cam}$ in the camera coordinate system. Subsequently, we transform these points into the ego-vehicle coordinate system using the extrinsic matrix $\mathbf{T}_{cam2ego}$:
\begin{equation}
    \mathbf{P}_{ego} = \mathbf{T}_{cam2ego} \cdot \left( \mathbf{D}(\mathbf{u}) \cdot \mathbf{K}^{-1} \tilde{\mathbf{u}} \right),
\end{equation}
where $\tilde{\mathbf{u}}$ is the homogeneous coordinate of pixel $\mathbf{u}$. We store the resulting point cloud along with its corresponding RGB values as $\mathcal{S}_{ego} = \{ (\mathbf{P}_{ego}, \mathbf{I}_{raw}(\mathbf{u})) \}$, which serves as the unified geometric representation for the subsequent virtual shift simulation.

\subsection{Virtual-Shift Strategy}
\label{sec:mask}

A core innovation of our method is how we utilize the $\mathcal{S}_{ego}$. Unlike prior works (\eg, FreeVS) that reconstruct the scene to synthesize a pseudo-target, we propose a Virtual-Shift strategy that circumvents the geometric errors inherent in monocular depth.

We first define a virtual camera pose $\mathbf{T}_{virt}$ by applying a spatial offset (\eg, a lateral shift of 1.0m) to the current ego pose. We project the ego-centric points $\mathcal{S}_{ego}$ onto this virtual image plane to generate a preliminary novel view, which we term the Shifts Image $\mathbf{I}_{shift}$.
\begin{equation}
    \mathbf{I}_{shift} = \text{Proj}(\mathcal{S}_{ego}, \mathbf{T}_{virt}, \mathbf{K}).
\end{equation}
Upon inspecting $\mathbf{I}_{shift}$, we observe significant artifacts: aside from the expected field-of-view loss, the image is riddled with ``black irregular curves" and voids inside object boundaries. These artifacts are caused by monocular depth errors and calibration inaccuracies, which are particularly pronounced around thin, repetitive structures such as lane dividers and fences.

Directly using $\mathbf{I}_{shift}$ as the input condition presents a fundamental problem: it represents a view shifted by 1.0m relative to the ground truth $\mathbf{I}_{raw}$. Training a model to map a 1.0m-shifted condition $\mathbf{I}_{shift}$ to the original pose $\mathbf{I}_{raw}$ forces it to learn an ill-posed warping function, confusing the model with the aforementioned projection artifacts.

To resolve this, we discard the pixel values of $\mathbf{I}_{shift}$ and retain only its validity information. We calculate a binary occlusion mask $\mathbf{M} \in \{0, 1\}^{H \times W}$, where $\mathbf{M}_{\mathbf{u}}=1$ indicates that the pixel at $\mathbf{u}$ is either occluded or projected out-of-view in the virtual frame $\mathbf{I}_{shift}$. We then apply this mask to the original raw image to obtain the Masked Image:
\begin{equation}
    \mathbf{I}_{masked} = \mathbf{I}_{raw} \odot (1 - \mathbf{M}).
\end{equation}
This operation effectively aligns the condition with the ground truth. By carving out the regions that would be lost in a virtual shift, we transform the task into recovering the missing pixels of $\mathbf{I}_{raw}$ itself. This allows us to achieve arbitrary shift distances during training without being constrained by the quality of the geometric proxy.

\subsection{Pseudo-3D Seam Synthesis}
\label{sec:seam}

While the Masked Image $\mathbf{I}_{masked}$ resolves the alignment issue, simply leaving the occluded regions as black voids introduces a new challenge: as the virtual shift distance increases (\eg, lateral shifts $>2$m), the mask $\mathbf{M}$ covers a significant portion of the image. Relying on the network to hallucinate such large-scale context from scratch often leads to severe spatial inconsistencies and temporal flickering.

To address this, we leverage the overlapping fields of view inherent in multi-camera autonomous driving setups. We observe that the ``blind spots" created by the virtual shift in the ego-view are often observable by adjacent cameras (\eg, a side camera). Therefore, we propose to fill these voids by projecting pixels from neighboring views, a process we term Pseudo-3D Seam Synthesis.

Specifically, given a neighboring image $\mathbf{I}_{nb}$ and its estimated monocular depth $\mathbf{D}_{nb}$, we warp its pixels to the current virtual pose $\mathbf{T}_{virt}$ to generate a candidate patch $\mathbf{I}_{warp}$. We then composite this patch into the masked region to form the final input condition $\mathbf{I}_{cond}$:
\begin{equation}
    \mathbf{I}_{cond} = \mathbf{I}_{masked} + (\mathbf{I}_{warp} \odot \mathbf{M}).
\end{equation}

It is important to emphasize that we do \textit{not} require $\mathbf{I}_{warp}$ to be perfectly aligned with the ground truth. In fact, $\mathbf{I}_{warp}$ naturally exhibits photometric discrepancies (due to different sensor exposures/ISPs) and geometric misalignments (due to monocular depth errors) relative to the target view. We explicitly retain these imperfections in the condition $\mathbf{I}_{cond}$. By training the network to map this ``imperfectly stitched" condition to the ``perfectly consistent" raw image $\mathbf{I}_{raw}$, we force the model to learn robust feature correction mechanisms. This effectively simulates the exposure changes and calibration noises encountered during real-world inference, transforming potential errors into a source of data augmentation.

\subsection{Generative Inpainting Network}
\label{sec:network}

Having constructed the self-supervised training pairs $\{\mathbf{I}_{cond}, \mathbf{I}_{raw}\}$, we employ a diffusion-based architecture to learn the reconstruction mapping. Our framework is built upon DiT~\cite{peebles2023dit}. 


\textbf{Latent Space Operation.} To reduce computational complexity, we first encode both the ground truth $\mathbf{I}_{raw}$ and the condition $\mathbf{I}_{cond}$ into a compressed latent space using a pre-trained VAE encoder $\mathcal{E}$, yielding latents $\mathbf{z}_{0} = \mathcal{E}(\mathbf{I}_{raw})$ and $\mathbf{c} = \mathcal{E}(\mathbf{I}_{cond})$. The condition latent $\mathbf{c}$ is then concatenated with the noisy latent $\mathbf{z}_{t}$ along the channel dimension to guide the denoising process.

\textbf{Training Objective.} We adopt a flow-matching objective~\cite{liu2022flows,lipman2024flowmatchingguidecode} for stable and efficient training. The network $\mathcal{F}_{\theta}$ takes the noisy latent $\mathbf{z}_{t}$, the time step $t$, and the condition $\mathbf{c}$ as input, and is trained to predict the velocity field $\mathbf{v}_{t}$ pointing towards the clean data $\mathbf{z}_{0}$. The optimization objective is defined as:
\begin{equation}
    \mathcal{L} = \mathbb{E}_{t, \mathbf{z}_0, \mathbf{c}} \left[ \| \mathcal{F}_{\theta}(\mathbf{z}_{t}, t, \mathbf{c}) - (\mathbf{z}_{1} - \mathbf{z}_{0}) \|^{2} \right],
\end{equation}
where $\mathbf{z}_{1}$ represents pure Gaussian noise. During inference, given a user-specified trajectory, we dynamically generate the mask and neighbor patches to construct $\mathbf{I}_{cond}$ on-the-fly, and use the trained network to synthesize the final photorealistic novel view.

\begin{table}[t]
\caption{Comparison with state-of-the-art methods on the nuScenes. \textbf{bold} and \underline{underlined}, respectively. "-" indicates that the metric was not reported.}
\renewcommand\arraystretch{1.1}
\resizebox{0.5\textwidth}{!}{
\begin{tabular}{l|cc|cc|cc}
\toprule
\multirow{2}{*}{Method} & \multicolumn{2}{c|}{Shift $\pm$ 1m} & \multicolumn{2}{c|}{Shift $\pm$ 2m} & \multicolumn{2}{c}{Shift $\pm$ 4m} \\
\cline{2-7}
                        & FID $\downarrow$         & FVD $\downarrow$          & FID $\downarrow$         & FVD $\downarrow$          & FID $\downarrow$         & FVD $\downarrow$          \\
\hline
PVG~\cite{chen2023periodic}                     & 48.15        & 246.74        & 60.44        & 356.23        & 84.50         & 501.16        \\
EmerNeRF~\cite{yang2023emernerf}                & 37.57        & 171.47        & 52.03        & 294.55        & 76.11        & 497.85        \\
StreetGaussain~\cite{huang2024s3gaussian}          & 32.12        & 153.45        & 43.24        & 256.91        & 67.44        & 429.98        \\
OmniRe~\cite{chen2025omnireomniurbanscene}                  & 31.48        & 152.01        & 43.31        & 254.52        & 67.36        & 428.20         \\
FreeVS~\cite{wang2024freevs}                  & 51.26        & 431.99        & 62.04        & 497.37        & 77.14        & 556.14        \\
DiST-4D~\cite{guo2025dist}                 & \textbf{10.12}        & \textbf{45.14}         & \underline{12.97}        & \underline{68.80}          & \underline{17.57}        & \underline{105.29}       \\
\hline \rowcolor{blue!15}
\textbf{Ours}                &  \underline{11.98}       &    \underline{46.27}     &    \textbf{12.48}    &  \textbf{65.36  }       &   \textbf{16.05 }    &       \textbf{98.78} \\
\bottomrule
\end{tabular}
}
\label{main}
\end{table}

\section{Experiments}

\subsection{Experimental Settings}
\label{sec:settings}

\subsubsection{Dataset and Metrics}
We evaluate our framework on the \textbf{nuScenes} dataset~\cite{caesar2020nuscenes}, a large-scale benchmark for autonomous driving consisting of 1000 driving scenes (700 for training, 150 for validation, and 150 for testing). Each scene is 20 seconds long and captured by a synchronized 6-camera array providing a $360^{\circ}$ field of view. We evaluate the quality of the generated RGB videos using the Fréchet Inception Distance (FID)~\cite{seitzer2020maximilian} for images and Fréchet Video Distance (FVD)~\cite{unterthiner2019fvd} for videos.

\subsubsection{Implementation Details.}
We initialize our framework using the pre-trained weights from Wan2.2~\cite{wan2025wan} to leverage its robust video generation priors. We freeze the VAE and all text encoder components, fine-tuning only the DiT~\cite{peebles2023dit} backbone. To accommodate our additional condition inputs, we modify the first convolution layer of the DiT to accept extra channels. These newly added parameters are initialized using Kaiming initialization~\cite{he2015delving}, while the original weights are inherited from Wan2.2. The model is trained on video sequences of 49 frames derived from the processed nuScenes dataset. We conduct training for 300 epochs on a cluster of 32 NVIDIA L20 GPUs (4 nodes $\times$ 8 GPUs). We employ a global batch size of 256 with gradient accumulation steps set to 8 to stabilize the optimization.
The training utilizes the AdamW optimizer~\cite{loshchilov2017decoupled} with hyperparameters $\beta_1=0.9$, $\beta_2=0.999$, and a weight decay of $0.01$ and BF16 mixed precision. The learning rate is set to $1 \times 10^{-4}$ and decays following a cosine schedule without warm-up. 



\begin{table}[t]
\centering
\caption{Ablation study of component effectiveness.}
\renewcommand\arraystretch{1.1}
\resizebox{0.5\textwidth}{!}{
\begin{tabular}{c|c|cc}
\toprule
 Virtual-Shift Strategy & Pseudo-3D Seam Synthesis & $FID$$\downarrow$ & $FVD$$\downarrow$\\
\hline
              &          & 21.76  & 56.80   \\
\checkmark    &          & 14.34  & 46.68  \\
\hline  \rowcolor{blue!15}
\checkmark    & \checkmark        & \textbf{11.98}  & \textbf{46.27}  \\
\bottomrule
\end{tabular}
}
\label{ablation}
\end{table}

\begin{figure}[b]
    \centering
    \includegraphics[width=0.99\linewidth]{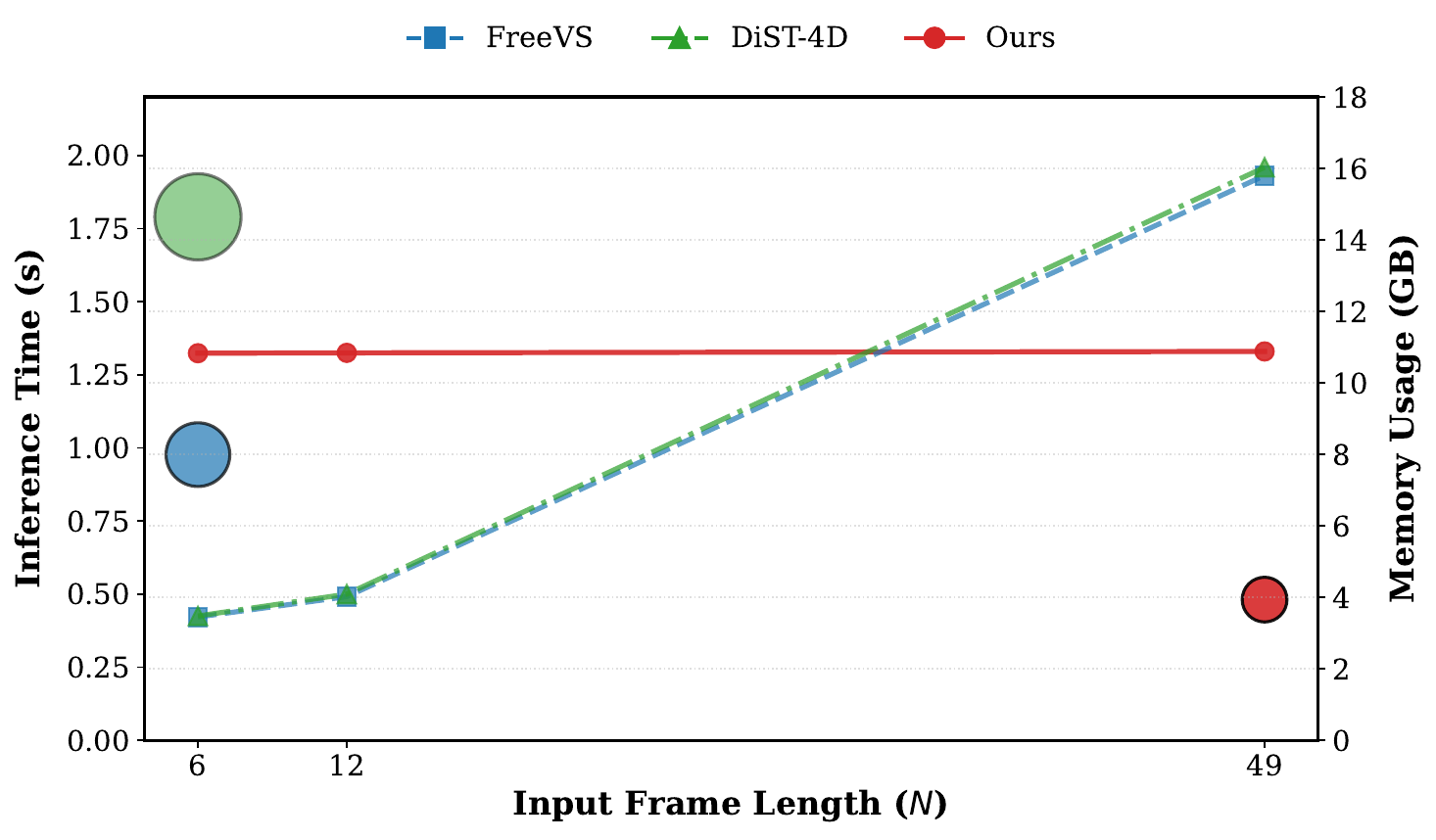}
    \caption{Comparison of computational efficiency. We profile memory usage and inference time under different input lengths $N$.}
    \label{fig:efficiency}
\end{figure}

\begin{figure*}[t]
    \centering
    \includegraphics[width=0.99\linewidth]{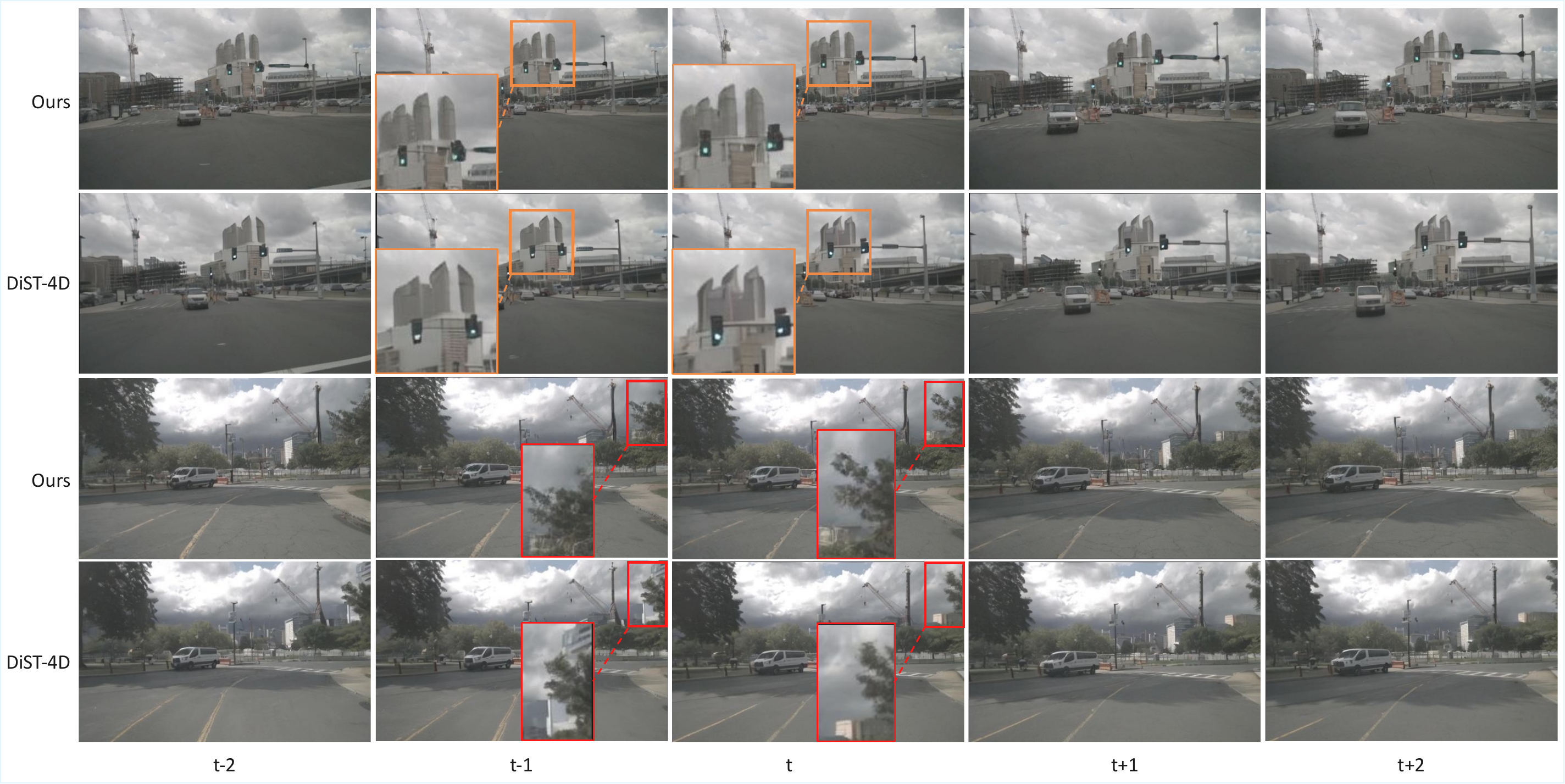}
    \caption{Qualitative comparison of temporal consistency. We visualize consecutive frames ($t-2$ to $t+2$) synthesized by DiST-4D~\cite{guo2025dist} and Ours. The orange and red boxes highlight specific regions of interest. DiST-4D exhibits severe temporal instability, with building structures distorting (top) and tree details flickering or vanishing (bottom) across frames. In contrast, Ours maintains remarkable geometric stability and visual coherence, preserving object details without artifacts.
    }
    \label{overview}
    \label{fig:qualitative}
\end{figure*}

\subsection{Main Results}
Table~\ref{main} presents the quantitative comparison on the NuScenes dataset across varying view shifts. First, our generative framework significantly outperforms reconstruction-based baselines (\eg, PVG, OmniRe). At large shifts of $\pm 4$m, VisionNVS reduces FID from $67.36$ (OmniRe) to $16.05$, demonstrating that generative priors are essential for synthesizing realistic details in unobserved regions where reconstruction methods fail.

Crucially, compared to the state-of-the-art DiST-4D~\cite{guo2025dist}, which relies on expensive LiDAR priors and 3D bounding boxes, our method operates in a strict \textbf{camera-only setting} yet achieves superior robustness. At near-field shifts ($\pm 1$m), our performance is highly competitive with DiST-4D (FID $11.98$ vs. $10.12$), validating that our self-supervised masking strategy effectively extracts precise geometry from monocular cues. Notably, at far-field shifts ($\pm 4$m), our method \textbf{surpasses} DiST-4D (FVD $98.78$ vs. $105.29$). This reversal highlights a fundamental advantage of our paradigm: while warping-based methods suffer from severe geometric degradation at large distances, our inpainting approach effectively maintains spatial consistency by treating large occlusions as a generative completion task, achieving SOTA performance with minimal sensor dependency.



\subsection{Ablation Studies}

To investigate the effectiveness of our core contributions, we conduct a component-wise ablation study on the NuScenes dataset, as summarized in Table~\ref{ablation}.
The baseline model (Row 1), which operates under the traditional extrapolation paradigm using warped pseudo-targets, yields suboptimal performance (FID 21.76). Incorporating the Virtual-Shift Strategy (Row 2) results in the most significant performance leap, improving FID by 7.42 and FVD by 10.12. This substantial gain validates our fundamental premise: reformulating NVS as an inpainting task—thereby utilizing artifact-free raw images as supervision—effectively eliminates the geometric distortions inherent in direct warping.
Furthermore, integrating Pseudo-3D Seam Synthesis (Row 3) provides additional improvements, further reducing FID to 11.98. By filling large occlusion voids with context from neighboring views, this module mitigates generative ambiguity in "blind spots," ensuring sharper texture details and better spatial consistency in the final output.

\begin{figure}[t]
    \centering
    \includegraphics[width=0.99\linewidth]{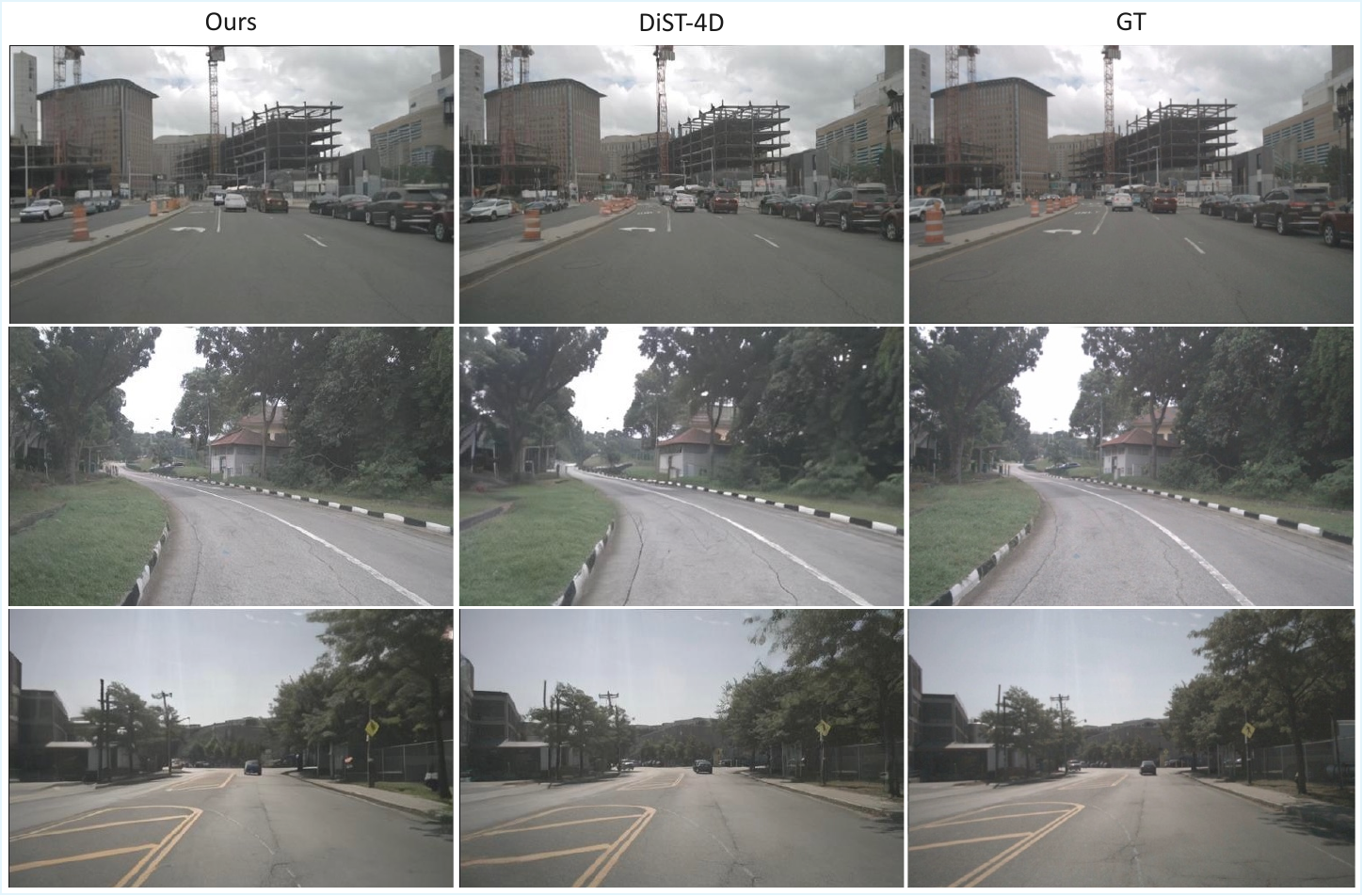}
    \caption{Visual comparison with state-of-the-art DiST-4D~\cite{guo2025dist}. Despite DiST-4D utilizing expensive LiDAR priors, dense metric depth supervision, and complex Cycle Consistency (SCC) strategies, our camera-only VisionNVS achieves comparable visual fidelity and geometric consistency without employing any of these additional tricks.}
    \label{fig:comparison_dist}
\end{figure}

\subsection{Efficiency Analysis}
\label{sec:efficiency}

We compare the computational efficiency against baselines by profiling inference latency and memory usage across varying input condition lengths ($N$). As shown in Figure~\ref{fig:efficiency}, baselines exhibit a linear growth in memory consumption as $N$ increases, reaching nearly 16GB at $N=49$. In contrast, thanks to our streaming inference paradigm, VisionNVS maintains a stable memory footprint regardless of input length. Remarkably, even when conditioning on a long-term context of 49 frames to maximize temporal consistency, our method achieves an inference time of 0.577s—roughly $\mathbf{3\times}$ faster than DiST-4D ($1.79$s), which only utilizes a short context of 6 frames. This demonstrates that our framework effectively scales to long-sequence modeling without the computational bottlenecks plaguing prior arts.

\subsection{Further Analysis}
\label{sec:qualitative}

\subsubsection{Qualitative results of temporal consistency}
Figure~\ref{fig:qualitative} compares the temporal coherence of synthesized video sequences. As observed, DiST-4D~\cite{guo2025dist} suffers from severe inter-frame jitter and temporal flickering, where background structures (\eg, buildings) distort unpredictably due to depth fluctuations. Furthermore, it fails to maintain semantic integrity, causing complex objects like trees to vanish or morph into noise during camera movement. In contrast, VisionNVS generates highly coherent sequences. Our model effectively suppresses the jarring "shaking" artifacts seen in the baseline and demonstrates superior object permanence. By leveraging the inpainting paradigm, our results remain geometrically stable, consistently reconstructing object details without sudden target loss or hallucination.

\subsubsection{Visual Comparison with SOTA and GT}
\label{sec:vis_sota}

To further evaluate visual fidelity, we compare VisionNVS with the DiST-4D~\cite{guo2025dist} in Figure~\ref{fig:comparison_dist}. As observed, both methods successfully reconstruct complex geometries, such as construction scaffolds (Row 1) and dense vegetation (Row 2). However, while DiST-4D relies on a heavy pipeline involving LiDAR priors, metric depth curation, and multi-stage SCC optimization, our method achieves comparable visual quality in a strict camera-only setting. This result validates that our self-supervised inpainting paradigm effectively resolves geometric ambiguities without expensive sensor dependencies, offering a significantly more scalable solution for autonomous driving simulation.

\section{Conclusion}
\label{sec:conclusion}

We propose VisionNVS, a camera-only framework that reformulates free-trajectory NVS as a self-supervised inpainting task. By leveraging the Virtual-Shift strategy and Pseudo-3D Seam Synthesis, we effectively bridge the supervision gap without relying on expensive LiDAR priors. Extensive experiments on nuScenes demonstrate that VisionNVS achieves state-of-the-art performance, surpassing LiDAR-dependent baselines in both visual fidelity and geometric consistency. By eliminating the dependency on active sensors, we hope VisionNVS paves the way for democratizing high-fidelity simulation, unlocking the full potential of massive internet-scale video data for autonomous driving research.

\bibliographystyle{IEEEbib}
\bibliography{icme}

\end{document}